# Z-Error Loss for Training Neural Networks
# A technical note

Guillaume Godin[1]

[1]Osmo Labs PCB New York, USA

Corresponding authors: guillaume@osmo.ai

# Abstract

Outliers introduce significant training challenges in neural networks by propagating erroneous gradients, which can degrade model performance and generalization. We propose the Z-Error Loss, a statistically principled approach that minimizes outlier influence during training by masking the contribution of data points identified as out-of-distribution within each batch. This method leverages batch-level statistics to automatically detect and exclude anomalous samples, allowing the model to focus its learning on the true underlying data structure. Our approach is robust, adaptive to data quality, and provides valuable diagnostics for data curation and cleaning.

# Scientific Contributions

- **Novel Outlier Mitigation:** Introduces Z-Error Loss, a batchwise masking approach that significantly reduces the effect of outliers on neural network training and model weights.

- **Automated Outlier Detection:** Enables automatic identification of anomalous or mislabeled samples, facilitating data cleaning and quality control.

- **Adaptive Thresholding:** Provides an annealing-inspired mechanism to dynamically adjust the inlier threshold during training, further improving model robustness.

- **Principled Threshold Selection:** Empirically infers the optimal classification cutoff based on the true distribution of network outputs, rather than relying on arbitrary defaults.

- **Versatile Application:** Extends to both regression and classification, and supports statistical modeling in both logit and probability space.

# Introduction

Normalization trick is a fundamental process of neural networks, mainly due to how loss propagation affects weight update across layers. We propose leveraging a normal (gaussian) distribution within the loss to automatically detect and reject outliers during the training. This offers the model a way to "forget" data errors instead of integrating them into the training model weights.

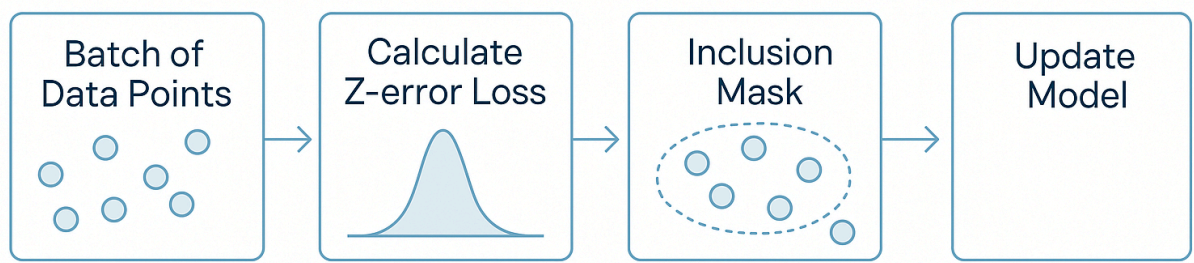

Figure 1: how to implement the Z-error loss ?

# Discussion

## Z-error loss for Regression

The Z-Error Loss is designed to minimize the effect of outliers during loss propagation. When training a model on batches of data, you can compute a batch-wise Z-error loss. That is, calculate the mean and standard deviation of the loss for the current batch (see figure 1, step 2). By creating an inclusion mask (see figure 1, step 3). For example, keeping only points with errors that fall within a validity range such as ±2σ, you can backpropagate errors only on inlier points, reducing the impact of true outliers (see figure 1, step 4). This method stabilizes training and prevents error propagation from outlier samples. In practice, this approach improves generalization, and enhances outlier detection. The Z-error loss offers an intuitive, data-driven way to exclude *in-situ* outliers without any prior knowledge: the model automatically learns its own inclusion/exclusion boundary.

Our method flags data points that require review, re-measurement, or correction. In many cases, such mistakes stem from unit conversion errors (Temperatures), float-point typos (12.3 vs 1.23), dyslexic typos (126 vs 162), data entry mistakes (845 vs 345) or data misalignment between values and objects (wrong Molecule attribution).

Using a similar method, It was found that the CIFAR database contains between 1-5% of images that are wrongly labeled after training the model (see Apply Z-error loss Classification).

During model training, the number of outliers typically changes as the model evolves: at the start, the model is poor at distinguishing inliers from outliers, but this stabilizes as training progresses. So the number of outliers reduces during training epochs until reaching a stable range value.

## Adaptive Sigma Threshold : an Annealing-Inspired Approach

We can also gradually decrease the sigma threshold during training, starting from a very large value (effectively including all points σ = 10) and reducing it toward a target value (e.g., σ = 2) as training progresses. This is similar to an annealing process, where the sigma threshold is synchronized with the total number of epochs. Early in training, the model uses a broad inclusion window, allowing it to learn from the entire dataset; as the model improves, the threshold narrows, focusing learning on inliers and excluding remaining outliers more aggressively (see get_sigma_threshold function).

## Z-error loss for Classification

The decision boundary in binary classification is, in theory, a mathematically infinitesimal point, rendering it fundamentally undecidable. In practice, this boundary is often ambiguous, especially for biological datasets where experimental noise and measurement uncertainty further blur the line between classes. For example, in biological assays, the experimental error can easily reach 1 log unit or more, making the classification around the decision boundary inherently uncertain. Labeling in such datasets often reflects the "best guess" rather than an absolute truth, so samples near the threshold are ambiguous not only for the model but also for human experts which is normally not taken into consideration during training of neural networks.

We propose to model the output of two class spaces using two independent Gaussian distributions: one for class 0, the other for class 1. This allows extending the Z-error masking approach to classification tasks. For each class, we compute the distribution of model outputs (e.g., logits or predicted probabilities) for samples of that class. Then, for each sample, we calculate a Z-score relative to its class distribution. By setting an inclusion threshold (e.g., within 2σ of the class mean), we focus training only on the inlier samples for each class. Outliers: samples whose predictions are far from the class mean, are excluded from the loss computation during that batch. We can set independent sigma for independent Gaussians and we can expand to cross category entropy loss. This extension enables the model to learn more robust and biologically meaningful decision boundaries, reduces the influence of ambiguous or mislabeled samples, and adapts seamlessly to both binary and multi-class classification scenarios.

## Infer the best cutoff threshold decision based on Z-error loss

An additional advantage of the Z-error loss framework is its capacity to guide the empirical determination of an optimal classification threshold, rather than relying on the conventional default (e.g., 0.5 for binary probability outputs). By leveraging the inlier distributions identified by Z-score masking, the model's output space for each class can be well-approximated by two separate Gaussian distributions (one for each class). The optimal threshold for classification can then be computed as the intersection point of these two Gaussian curves, which represents the value at which the likelihood of a sample belonging to either class is maximized and equally probable. This data-driven approach allows for a more accurate and adaptive separation between classes, especially in cases where the underlying distributions are asymmetric or affected by class imbalance, label noise, or biological ambiguity. The threshold may be updated dynamically throughout training or established post hoc on a validation set, thereby providing a principled and robust alternative to the arbitrary 0.5 cutoff. This further enhances the model's capacity to generalize and make reliable predictions in challenging, real-world scenarios.

## Gaussian vs. Skew-Normal Distributions: Before vs. After the Sigmoid Nonlinearity

We prefer to apply our statistical modeling and Z-error masking in the logit space, as it represents the native probability domain of neural networks and preserves the linearity of model outputs. In this space, output distributions for each class are typically well-approximated by Gaussian functions. However, it is also possible to apply the method after the sigmoid activation, working directly with predicted probabilities. Since the sigmoid function is nonlinear and compresses the range of outputs to [0, 1], it distorts the underlying distribution and can introduce skewness, particularly for confident models. In such cases, a skew-normal distribution provides a more appropriate fit than a standard Gaussian. Thus, while both approaches are valid, the choice of modeling distribution should reflect the space in which the statistical analysis is performed.

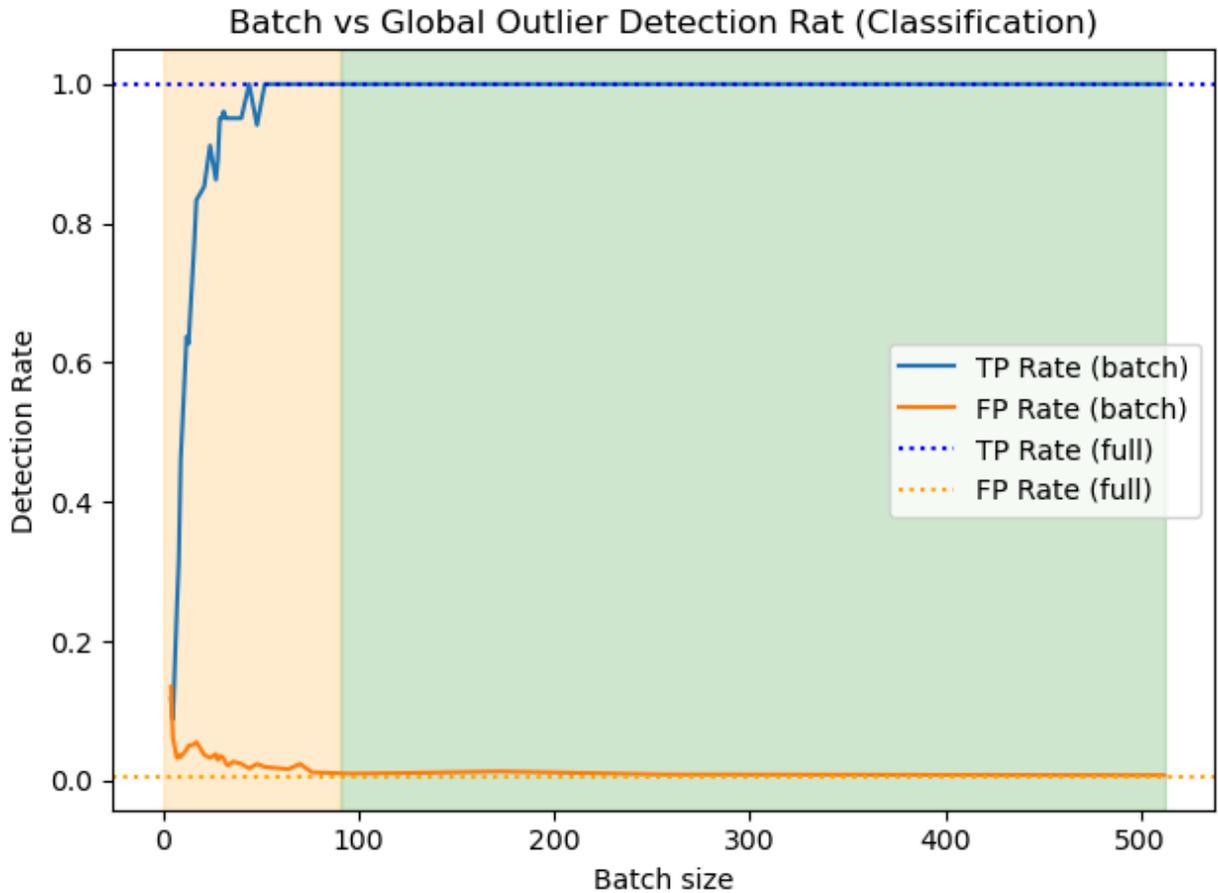

Figure 2 example with 10% synthetic outliers on both balanced classes detected using batch and full dataset. At low batch size the detection score with the batch method is less accurate. 1.5σ.

## Limitations

As this method operates in a statistical space, its reliability and effectiveness improve with larger batch sizes: ideally greater than 96, and preferably 256 or more (see figures 2,3). Smaller batches may not sufficiently capture the underlying distributions of the data, leading to less stable threshold estimation and outlier detection. This limitation applies equally to the selection of the optimal cutoff threshold; to obtain a robust and generalizable threshold value, it is recommended but not obligatory to aggregate model outputs over the entire dataset or a large validation set and compute the threshold in a post-processing step (see figures 2,4).

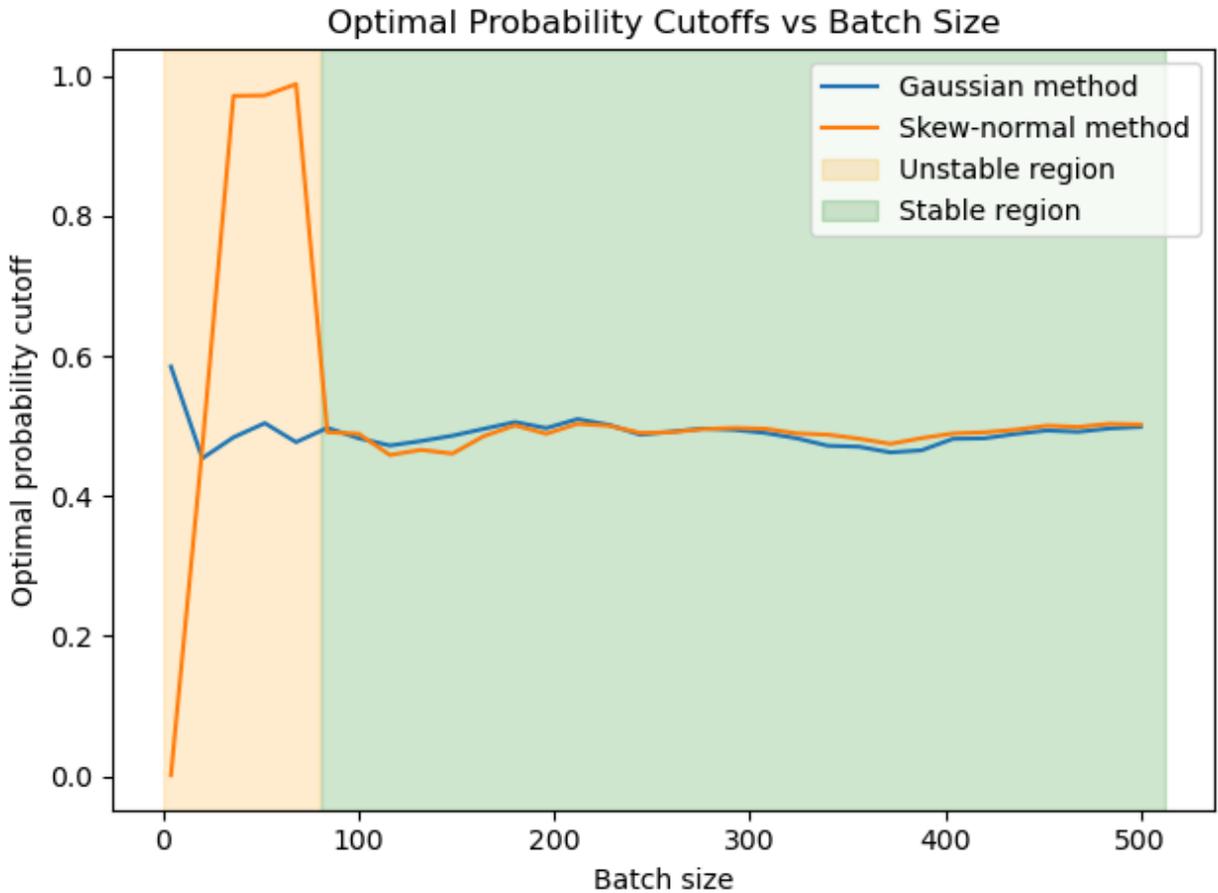

Figure 3 Statistical stable regions of Gaussian versus Skew normal distributions evolution for two uniform balanced synthetic datasets based on batch size.

# Conclusion

In this technical work, we introduce the Z-Error Loss, a simple yet powerful extension to classical loss functions in neural network training. By leveraging batchwise statistical analysis, our method adaptively masks the influence of outliers and ambiguous samples, thereby enhancing both model robustness and interpretability. The Z-Error Loss requires only a single tunable parameter (the inlier threshold) making it straightforward to implement and calibrate for diverse datasets and tasks. Beyond improved generalization and reduced sensitivity to data noise, Z-Error Loss also provides actionable insights for data quality assessment and model diagnostics. We recommend its adoption as a general-purpose tool for training robust neural networks, especially in domains where data uncertainty or mislabeling are prevalent.

# Competing Interests and Consent for publication

The author declares that he has no competing interests. The author has read and agreed to the published version of the manuscript.

# Acknowledgement

I would like to thank Dr. Ruud Van Deursen for the many hours he devoted to teaching me the importance of statistical analysis in neural networks. His guidance and insights were instrumental in shaping the foundations of this work.

# Code Section

## 1. Z-error loss for Regression

```python
class ZMSELoss(nn.Module):
    def __init__(self, threshold=2.0):
        """Z-MSE Loss that ignores outliers beyond a Z-score threshold."""
        super().__init__()
        self.threshold = threshold  # 2-sigma threshold (adjustable) recommended
    def forward(self, predictions, targets):
        """
        Compute the MSE loss while ignoring outliers beyond `self.threshold` sigma.
        Args:
            predictions (torch.Tensor): Model output of shape [batch_size, 1]
            targets (torch.Tensor): Ground truth of shape [batch_size, 1]
        Returns:
            torch.Tensor: Z-MSE Loss
        """
        if targets.ndim == 1:
            targets = targets.unsqueeze(-1)
        if predictions.ndim == 1:
            predictions = predictions.unsqueeze(-1)
        mean = torch.mean(targets)
        std = torch.std(targets, unbiased=True)
        z_scores = (targets - mean) / (std + 1e-8)
        mask = (z_scores.abs() <= self.threshold).float()  # Mask for inlier values
        mse_loss = F.mse_loss(predictions, targets, reduction="none")
        masked_loss = mse_loss * mask
        valid_count = mask.sum()
        return masked_loss.sum() / (valid_count + 1e-8)  # Avoid division by zero
```

## 2. Adaptive Sigma Threshold : an Annealing-Inspired Approach

```python
def get_sigma_threshold(epoch, max_epochs, start_sigma=100.0, end_sigma=2.0):
    # Linearly anneal sigma threshold
    progress = epoch / max_epochs
    return start_sigma + (end_sigma - start_sigma) * progress
```

## 3. Z-error loss for Classification

```python
import torch
import torch.nn as nn
import torch.nn.functional as F

class ZErrorBCEWithLogitsLoss(nn.Module):
    def __init__(self, threshold=2.0):
        """
        Z-error masked binary cross-entropy loss.
        Only inlier samples (by Z-score in their class's logit distribution) contribute to the loss.
        """
        super().__init__()
        self.threshold = threshold

    def forward(self, logits, labels):
        """
        logits: [batch_size, 1] - model outputs (before sigmoid)
        labels: [batch_size, 1] or [batch_size] - binary labels (0 or 1)
        """
        if logits.ndim == 1:
            logits = logits.unsqueeze(-1)
        if labels.ndim == 1:
            labels = labels.unsqueeze(-1)
        mask = torch.zeros_like(labels, dtype=torch.float32)
        bce_loss = F.binary_cross_entropy_with_logits(logits, labels.float(), reduction="none")
        # Apply Z-error logic separately for each class
        for cls in [0, 1]:
            idx = (labels.squeeze(-1) == cls).nonzero(as_tuple=True)[0]
            if len(idx) == 0:
                continue
            class_logits = logits[idx]
            mean = class_logits.mean()
            std = class_logits.std(unbiased=True)
            if std < 1e-8:  # avoid zero division
```

```python
            std = 1.0
        z_scores = ((class_logits - mean) / std).abs()
        class_mask = (z_scores <= self.threshold).float()
        mask[idx] = class_mask
    masked_loss = bce_loss * mask
    valid_count = mask.sum()
    return masked_loss.sum() / (valid_count + 1e-8)  # Prevent division by zero

# Example usage
if __name__ == "__main__":
    # Fake batch
    logits = torch.tensor([2.5, 0.2, -1.1, 1.3, -2.0, 3.0], dtype=torch.float32)  # model raw outputs
    labels = torch.tensor([1, 1, 0, 1, 0, 1], dtype=torch.float32)
    loss_fn = ZErrorBCEWithLogitsLoss(threshold=1.1)
    loss = loss_fn(logits, labels)
    print(f"Z-error-masked BCE loss @th 1.1 : {loss.item():.3f}" )
    loss_fn = ZErrorBCEWithLogitsLoss(threshold=1.3)
    loss = loss_fn(logits, labels)
    print(f"Z-error-masked BCE loss @th 1.3 : {loss.item():.3f}")
```

## 4. Infer the best cutoff threshold decision based on Z-error loss

```python
import torch
import numpy as np
from scipy.stats import norm, skewnorm
from scipy.optimize import brentq

def z_error_inlier_mask(logits, labels, threshold=2.0):
    mask = torch.zeros_like(logits, dtype=torch.bool)
    for cls in [0, 1]:
        idx = (labels == cls).nonzero(as_tuple=True)[0]
        if len(idx) == 0:
            continue
        class_logits = logits[idx]
        mean = class_logits.mean()
        std = class_logits.std(unbiased=True)
        if std < 1e-8: std = 1.0
        z_scores = ((class_logits - mean) / std).abs()
        class_mask = (z_scores <= threshold)
        mask[idx] = class_mask
    return mask

def optimal_logit_cutoff_from_gaussians(logits, labels, z_threshold=2.0):
    mask = z_error_inlier_mask(logits, labels, threshold=z_threshold)
```

```python
        logits_inlier = logits[mask]
        labels_inlier = labels[mask]
        logits_0 = logits_inlier[labels_inlier == 0].cpu().numpy()
        logits_1 = logits_inlier[labels_inlier == 1].cpu().numpy()
        if len(logits_0) < 2 or len(logits_1) < 2:
            raise ValueError("Not enough inlier points in each class to fit Gaussians.")
        mu0, std0 = logits_0.mean(), logits_0.std()
        mu1, std1 = logits_1.mean(), logits_1.std()
        def find_gauss_intersection(mu0, std0, mu1, std1):
            a = 1/(2*std0**2) - 1/(2*std1**2)
            b = mu1/(std1**2) - mu0/(std0**2)
            c = (mu0**2)/(2*std0**2) - (mu1**2)/(2*std1**2) - np.log(std1/std0)
            roots = np.roots([a, b, c])
            return np.real(roots[np.isreal(roots)])
        cutoffs = find_gauss_intersection(mu0, std0, mu1, std1)
        chosen_cutoff = cutoffs[np.argmin(np.abs(cutoffs - (mu0 + mu1)/2))]
        sigmoid = lambda x: 1 / (1 + np.exp(-x))
        return chosen_cutoff, sigmoid(chosen_cutoff)

def optimal_prob_cutoff_from_skewnorm(logits, labels, z_threshold=2.0):
    probs = torch.sigmoid(logits)
    mask = z_error_inlier_mask(probs, labels, threshold=z_threshold)
    probs_inlier = probs[mask]
    labels_inlier = labels[mask]
    probs_0 = probs_inlier[labels_inlier == 0].cpu().numpy()
    probs_1 = probs_inlier[labels_inlier == 1].cpu().numpy()
    if len(probs_0) < 2 or len(probs_1) < 2:
        raise ValueError("Not enough inlier points in each class to fit SkewNormals.")
    # Fit skewnorms
    skew0 = skewnorm.fit(probs_0)
    skew1 = skewnorm.fit(probs_1)
    pdf0 = lambda x: skewnorm.pdf(x, *skew0)
    pdf1 = lambda x: skewnorm.pdf(x, *skew1)
    # Find intersection numerically in [0, 1]
    try:
        intersection = brentq(lambda x: pdf0(x) - pdf1(x), 0.001, 0.999)
    except ValueError:
        # If no intersection in range, fallback to midpoint
        intersection = (probs_0.mean() + probs_1.mean()) / 2
    return intersection

if __name__ == "__main__":
    n=1000
    np.random.seed(42)  # For reproducibility
```

```python
# Generate 100 logits for class 0 (centered at -2, some spread)
logits_0 = np.random.normal(loc=-2.0, scale=1.0, size=n)
# Generate 100 logits for class 1 (centered at +2, some spread)
logits_1 = np.random.normal(loc=2.0, scale=1.0, size=n)
# Concatenate and shuffle
logits_all = np.concatenate([logits_0, logits_1])
labels_all = np.array([0]*n + [1]*n)
indices = np.arange(2*n)
np.random.shuffle(indices)
logits_all = logits_all[indices]
labels_all = labels_all[indices]
logits = torch.tensor(logits_all, dtype=torch.float32)
labels = torch.tensor(labels_all, dtype=torch.float32)
z_threshold = 2.0  # Or anneal over epochs!
# Gaussian method
logit_cutoff, prob_cutoff_gauss = optimal_logit_cutoff_from_gaussians(logits, labels, z_threshold)
print(f"Optimal logit cutoff (Gaussian): {logit_cutoff:.3f}")
print(f"Optimal probability cutoff (Gaussian): {prob_cutoff_gauss:.3f}")
# Skew-normal method
prob_cutoff_skew = optimal_prob_cutoff_from_skewnorm(logits, labels, z_threshold)
print(f"Optimal probability cutoff (SkewNorm): {prob_cutoff_skew:.3f}")
```

# Experiments on batch outliers detection using synthetic data

For those plots (figures 2 and 4) we consider adding easy outliers points (10%) into the dataset with a margin offset to be sure, such points are out of the distribution for both regression and classification. We are only interested to study the batch size impact on the method here so synthetic data is enough to understand the phenomena.

We take for both a $1.5\sigma$ threshold. Basically statistical values are not pertinent at low batch size even if the data is clearly easy to separate (as synthetic). With a $2\sigma$ the batch size needs to be larger as we integrate more noise in the loss (see figures 5 & 6).

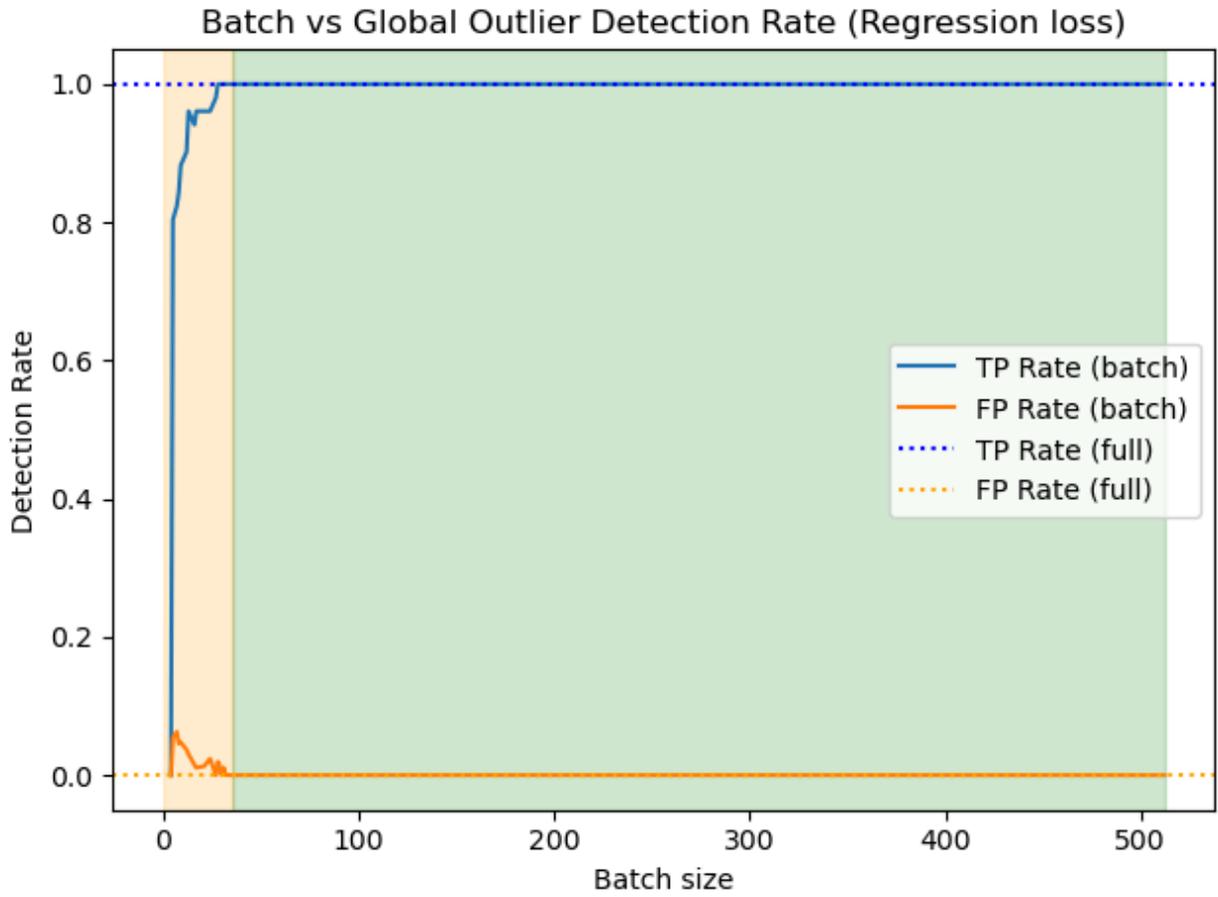

Figure 4 example with 10% synthetic outliers on regression points detected using batch and full dataset. At low batch size the detection score with the batch method is less accurate, 1.5σ.

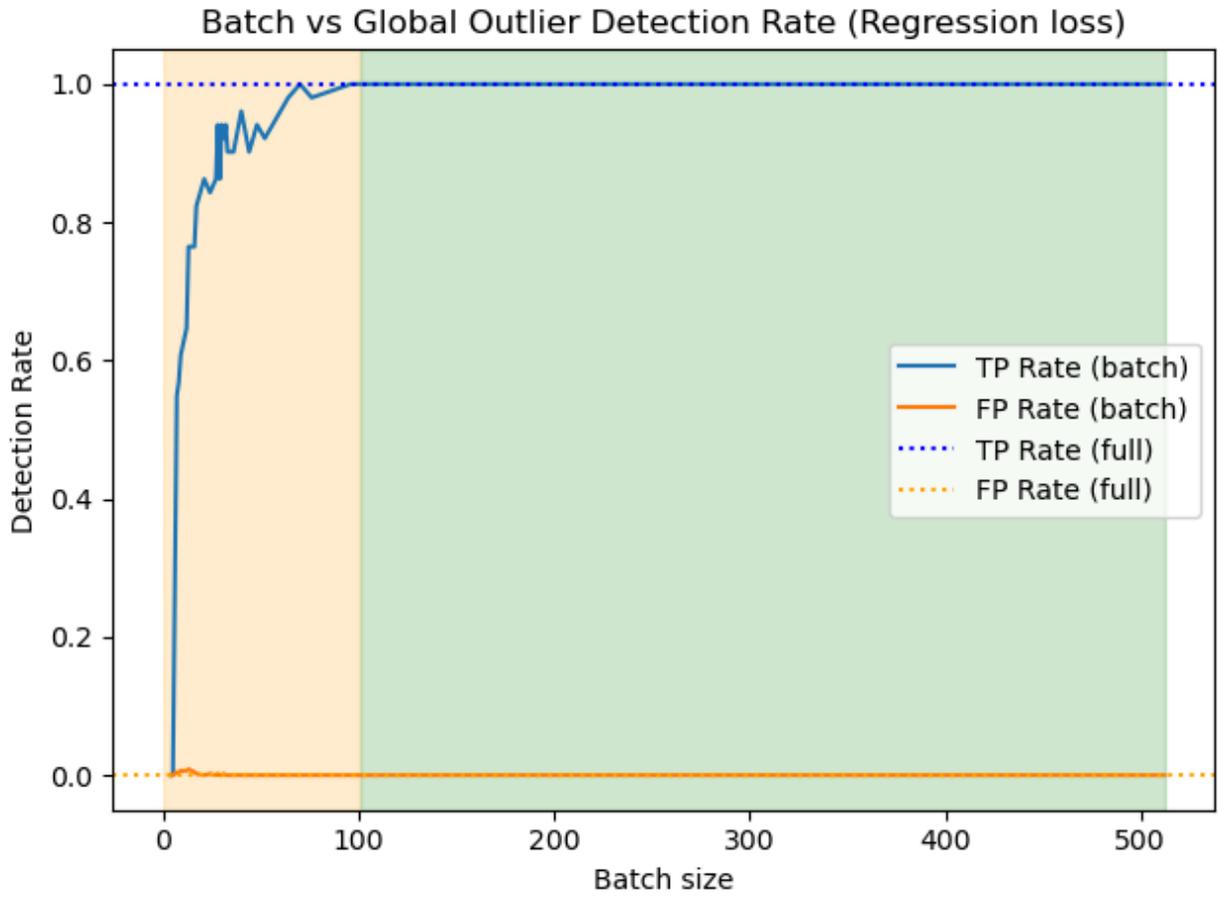

Figure 5 example with 10% synthetic outliers on regression points detected using batch and full dataset. At low batch size the detection score with the batch method is less accurate, 2σ.

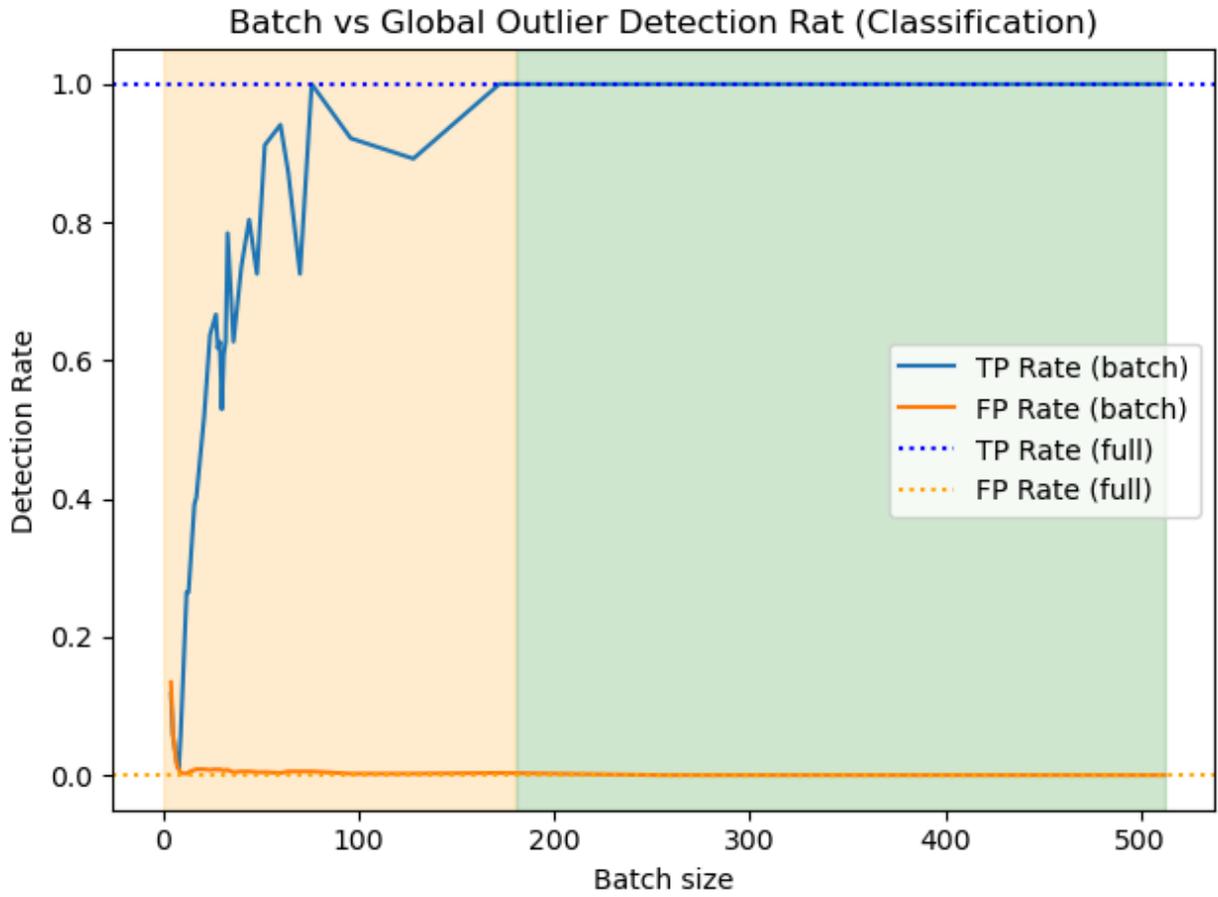

Figure 6 example with 10% synthetic outliers on both balanced classes detected using batch and full dataset. At low batch size the detection score with the batch method is less accurate. 2σ.